\documentclass{article}

\usepackage{microtype}
\usepackage{graphicx}
\usepackage{subfigure}
\usepackage{booktabs} 
\usepackage[ruled]{algorithm2e}
\usepackage{hyperref}

\usepackage[accepted]{sysml2019}

\sysmltitlerunning{}

\newcommand{\astfootnote}[1]{
\let\oldthefootnote=\thefootnote
\setcounter{footnote}{1}
\renewcommand{\thefootnote}{\fnsymbol{footnote}}
\footnote{#1}
\let\thefootnote=\oldthefootnote
}

\begin{document}

\twocolumn[
\sysmltitle{SkyNet: a Hardware-Efficient Method for Object Detection and Tracking on Embedded Systems}
\vspace{-15pt}



\begin{sysmlauthorlist}
\sysmlauthor{Xiaofan Zhang}{ui}
\sysmlauthor{Haoming Lu}{ui}
\sysmlauthor{Cong Hao}{ui}
\sysmlauthor{Jiachen Li}{ui}
\sysmlauthor{Bowen Cheng}{ui}
\sysmlauthor{Yuhong Li}{ui}
\sysmlauthor{Kyle Rupnow}{iot}
\sysmlauthor{Jinjun Xiong}{ibm,ui}
\sysmlauthor{Thomas Huang}{ui}
\sysmlauthor{Honghui Shi}{ibm,ui}
\sysmlauthor{Wen-mei Hwu}{ui}
\sysmlauthor{Deming Chen}{ui,iot}
\end{sysmlauthorlist}

\sysmlaffiliation{ui}{IBM-Illinois Center for Cognitive Computing Systems Research (C3SR), University of Illinois at Urbana-Champaign, USA}
\sysmlaffiliation{iot}{Inspirit IoT, Inc, USA}
\sysmlaffiliation{ibm}{IBM Research, USA}

\sysmlcorrespondingauthor{Xiaofan Zhang}{xiaofan3@illinois.edu \vspace{-8pt}}

\sysmlkeywords{Machine Learning, on-device intelligence, hardware-efficient computing}

\vskip 0.12in

\begin{abstract}
Object detection and tracking are challenging tasks for resource-constrained embedded systems. While these tasks are among the most compute-intensive tasks from the artificial intelligence domain, they are only allowed to use limited computation and memory resources on embedded devices. In the meanwhile, such resource-constrained implementations are often required to satisfy additional demanding requirements such as real-time response, high-throughput performance, and reliable inference accuracy. To overcome these challenges, we propose SkyNet, a hardware-efficient neural network to deliver the state-of-the-art detection accuracy and speed for embedded systems. Instead of following the common top-down flow for compact DNN (Deep Neural Network) design, SkyNet provides a bottom-up DNN design approach with comprehensive understanding of the hardware constraints at the very beginning to deliver hardware-efficient DNNs. The effectiveness of SkyNet is demonstrated by winning the competitive System Design Contest for low power object detection in the 56th IEEE/ACM Design Automation Conference (DAC-SDC), where our SkyNet significantly outperforms all other 100+ competitors: it delivers 0.731 Intersection over Union (IoU) and 67.33 frames per second (FPS) on a TX2 embedded GPU; and 0.716 IoU and 25.05 FPS on an Ultra96 embedded FPGA. The evaluation of SkyNet is also extended to GOT-10K, a recent large-scale high-diversity benchmark for generic object tracking in the wild. For state-of-the-art object trackers SiamRPN++ and SiamMask, where ResNet-50 is employed as the backbone, implementations using our SkyNet as the backbone DNN are 1.60X and 1.73X faster with better or similar accuracy when running on a 1080Ti GPU, and 37.20X smaller in terms of parameter size for significantly better memory and storage footprint.


\vspace{-10pt}
\end{abstract}
]



\printAffiliationsAndNotice{}  

\vspace{-5pt}
\section[title]{Introduction
}
\vspace{-5pt}
\label{sec:introduction}

Edge AI applications require not only high inference accuracy from DNNs, but also 
aggressive inference speed, throughput, and energy efficiency to meet real-life demands. These applications rely on hardware-efficient DNN designs when they are deployed 
into embedded systems with extremely limited computation and memory resources. Recently, we have seen intensive studies on DNN accelerators in hardware, which attempt to take advantage of different hardware design styles, such as GPUs, FPGAs and ASICs, to improve the speed and efficiency of DNN inference and training processes \cite{qiu2016going,isscc_2016_chen_eyeriss,zhang2017high,jouppi2017datacenter,franklin2017nvidia,zhang2018dnnbuilder,li2019implementing,chen2019cloud}.

Although hardware accelerators can be helpful, they are still limited by available resources to handle varied real-life applications, especially for embedded systems since most DNNs are not originally designed to be hardware-efficient. As a result,
developers started to focus their optimization efforts on 
the software side, to compress DNNs for less complexities, lowering computation demands and memory footprints. Recent researches have demonstrated the possibility of using low bit-width data to represent original floating-point parameters, such as using binary and ternary networks \cite{courbariaux2016binarized,rastegari2016xnor,li2016ternary,tschannen2018strassennets,wang2018design,gope2019ternary}. These solutions are intended to replace the hardware-intensive floating-point multiplications by logical operations, so that DNNs can be more efficient on hardware platforms.

\begin{figure*}
\centering
\vspace{-5pt}
\includegraphics[width=0.67\textwidth]{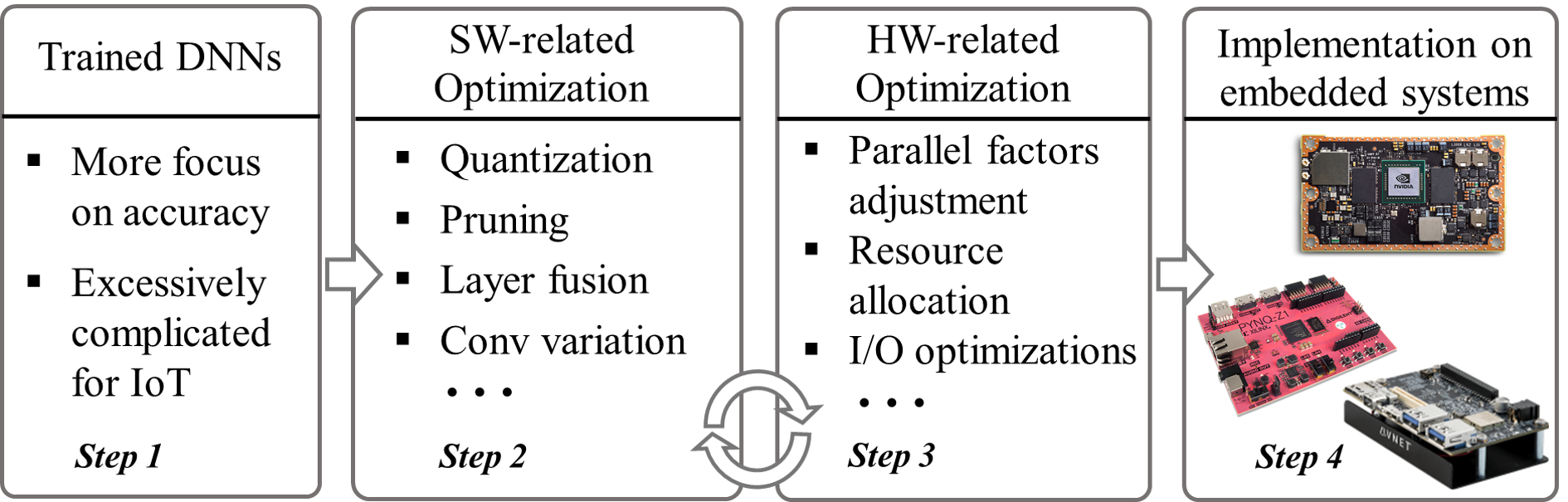}
\vspace{-10pt}
\caption{A top-down design flow for hardware-efficient DNN deployment on resource-constrained embedded systems. Challenges appear between \textit{step 2} and \textit{3} where iterative explorations are necessary to balance DNN accuracy and performance on targeted devices.}
\vspace{-10pt}
\label{fig:com_flow}
\end{figure*}

Researchers also investigate the network pruning strategies to reduce the redundancy of DNN structures \cite{han2015learning,han2016deep,luo2017thinet}. According to the published pruning strategies, the relatively less important connections between DNN layers are discarded and network retraining is then performed to regain accuracy. Significant reductions can be achieved on the classic DNNs, such as AlexNet \cite{krizhevsky2012imagenet} and VGG-16 \cite{simonyan2014very}. Since the major benefit of network compression comes from the fully-connected (FC) layers, to continuously have effective pruning results for latter DNNs (e.g., GoogleNet \cite{szegedy2015going} and ResNet \cite{he2016deep}) with reduced FC layers, more sophisticated algorithms are required to be integrated in network pruning, such as using evolutionary algorithms \cite{dai2019nest}, alternating direction method of multipliers (ADMM) \cite{ren2019admm}, and iterative pruning \cite{ding2018auto}.

As most of the computations happen inside the convolutional (Conv) layers, previous works also attempt to reduce the computation complexity by using depth-wise separable Conv layers for image classification and ubiquitous keyword-spotting applications \cite{howard2017mobilenets,zhang2017hello}. The depth-wise separable structure can 
{\color{black} effectively}
reduce the number of operations and provide more compact DNN designs for resource-constrained hardware. To further improve the DNN deployment on hardware, layer fusion is proposed in \cite{alwani2016fused} to minimize data movements between on-chip and off-chip memory.

In general, a {\color{black} traditional} design process 
{\color{black} for} hardware-efficient DNNs can be summarized in Figure \ref{fig:com_flow} with the adoption of above-mentioned technologies. It is a top-down design flow which starts from \textit{step 1}: to select a reference DNN 
{\color{black} by concentrating} 
on accuracy. 
Such DNNs are {\color{black} typically too} 
complicated for targeted embedded systems
{\color{black} and need to}
be compressed using software and hardware optimizations in \textit{step 2} and \textit{3}, respectively. Since software compression and hardware implementation are typically carried out in two
separate steps, \textit{step 2} and \textit{3} are usually performed in an iterative manner to balance DNN accuracy and hardware performance on targeted devices. Network retraining is also required to regain accuracy after compression before \textit{step 4}. Because of the iterative nature of the process, it is very challenging to cover both inference accuracy in software and deployment efficiency in hardware. 


In this paper, we address the hardware-efficient DNN design problem by proposing SkyNet, which is designed following a bottom-up DNN design approach with comprehensive awareness of hardware constraints. 
The main contributions of this paper are summarized as follows:
\begin{itemize}
\vspace{-5pt}
\setlength{\itemsep}{-3pt}
    \item We 
    {\color{black} survey} the latest low power object detectors for embedded systems and 
    {\color{black} identify} the potential obstacles of using top-down DNN design flows, which may prevent improved DNN accuracy and hardware efficiency.
    \item We propose a bottom-up design strategy of hardware-efficient DNNs for both embedded GPU and embedded FPGA; using such a design method, we propose SkyNet, which has comprehensive awareness of hardware limitations to overcome the challenges of top-down design flow. 
    \item We demonstrate SkyNet in DAC-SDC'19 using both TX2 GPU and Ultra96 FPGA with the stat-of-the-art accuracy. SkyNet achieved the highest overall score regarding accuracy, throughput, and energy-efficiency, and won the first place winner award for both GPU and FPGA tracks.
    \item We extend SkyNet for object tracking. By using SkyNet as the backbone DNN, SiamRPN++ and SiamMask obtain 1.60X and 1.73X speedup with better or similar accuracy, and 37.20X smaller parameter size compared to using the original ResNet-50 backbone when running on a 1080Ti GPU.
    \vspace{-5pt}
\end{itemize}

\vspace{-5pt}
\section{Related Work}
\vspace{-5pt}
\begin{table*}[t]
\vspace{-10pt}
\caption{DAC-SDC winning entries from both GPU and FPGA tracks. They follow a top-down approach, from choosing reference DNNs to applying optimization strategies on software and hardware sides, so that they compress DNNs with improved hardware efficiency. Optimizations include: \textcircled{\tiny{1}} input resizing, \textcircled{\tiny{2}} network pruning, \textcircled{\tiny{3}} data quantization, and \textcircled{\tiny{4}} TensorRT \cite{vanholder2016efficient} on software, and \textcircled{\tiny{5}} CPU-FPGA task partition, \textcircled{\tiny{6}} double-pumped DSP, \textcircled{\tiny{7}} fine-grained pipeline, \textcircled{\tiny{8}} clock gating, and \textcircled{\tiny{9}} multithreading on hardware.}
\vspace{-5pt}
\label{tab:dacsdc_summary}
\vskip 0.15in
\begin{center}
\newcommand{\tabincell}[2]{\begin{tabular}{@{}#1@{}}#2\end{tabular}}
\begin{sc}
\footnotesize
\begin{tabular}{c|c|c|c|c}
\toprule
\textbf{  Rank  } & \textbf{GPU-Track} & \textbf{Reference DNN}& 
\multicolumn{2}{c}{\textbf{  Optimizations}}\\
\midrule
'19 2nd & \tabincell{c}{Thinker \cite{thinker}}& ShuffleNet + RetinaNet &
\textcircled{\tiny{1}} \textcircled{\tiny{2}} \textcircled{\tiny{3}} & \textcircled{\tiny{9}} \\ 
'19 3rd & \tabincell{c}{DeepZS \cite{DeepZS}}& Tiny YOLO &
Not clear & \textcircled{\tiny{9}} \\ 
'18 1st & \tabincell{c}{ICT-CAS \cite{ICT-CAS}} & Tiny YOLO &
\textcircled{\tiny{1}} \textcircled{\tiny{2}} \textcircled{\tiny{3}} \textcircled{\tiny{4}} & Not clear \\ 
'18 2nd & \tabincell{c}{DeepZ \cite{DeepZ}}& Tiny YOLO & 
Not clear & \textcircled{\tiny{9}} \\ 
'18 3rd & \tabincell{c}{SDU-Legend \cite{SDU-Legend}}& YOLOv2 &
\textcircled{\tiny{1}} \textcircled{\tiny{2}} \textcircled{\tiny{3}} & \textcircled{\tiny{9}}\\ 
\midrule
\textbf{Rank} & \textbf{FPGA-Track} & \textbf{Reference DNN} &
\multicolumn{2}{c}{\textbf{  Optimizations}} \\
\midrule
'19 2nd & \tabincell{c}{XJTU Tripler \cite{XJTU}}& ShuffleNetV2 + YOLO & 
\textcircled{\tiny{2}} \textcircled{\tiny{3}} & \textcircled{\tiny{5}} \textcircled{\tiny{6}} \textcircled{\tiny{8}}\\ 
'19 3rd & \tabincell{c}{SystemsETHZ \cite{systemsETHZ19}}& SqueezeNet + YOLO& 
\textcircled{\tiny{1}} \textcircled{\tiny{2}} \textcircled{\tiny{3}} & \textcircled{\tiny{7}}\\ 
'18 1st & \tabincell{c}{TGIIF \cite{TGIIF}} & SSD & 
\textcircled{\tiny{1}} \textcircled{\tiny{2}} \textcircled{\tiny{3}} & \textcircled{\tiny{5}} \textcircled{\tiny{6}}\\
'18 2nd & \tabincell{c}{SystemsETHZ \cite{systemsETHZ}} & SqueezeNet + YOLO& 
\textcircled{\tiny{1}} \textcircled{\tiny{2}} \textcircled{\tiny{3}} & \textcircled{\tiny{7}}\\
'18 3rd & \tabincell{c}{iSmart2 \cite{iSmart2}} & MobileNet + YOLO&
\textcircled{\tiny{1}} \textcircled{\tiny{2}} \textcircled{\tiny{3}} & \textcircled{\tiny{5}} \textcircled{\tiny{7}}  \\ 
\bottomrule
\end{tabular}
\end{sc}
\end{center}
\vskip -0.15in
\end{table*}
Recent state-of-the-art object detectors feature DNN backbones to extract input features.
{\color{black} Researchers} initially propose a two-stage approach which first outputs multiple region proposals for object candidates and then generates more accurate regions with corresponding class labels \cite{dai2016r,lin2017feature,he2017mask, cheng2018decoupled,cheng2018revisiting,li2019scaleaware}. To improve the detection speed, some one-stage approaches are proposed to simultaneously regress object locations and classes \cite{sermanet2014overfeat,redmon2016you,liu2016ssd,lin2017focal,shen2019improving,tian2019fcos}. Object tracking also relies on the features extracted from DNN backbones, and we have seen recent Siamese network based trackers formulate trackers as feature between the exemplar image and search region \cite{Tao_2016,Valmadre_2017,wang2018learning,li2018siamrpn++,wang2019fast}. 
These emerging methods make real-time object detection and tracking possible using desktop GPUs but still need aggressive compression before deploying onto embedded systems.

\vspace{-5pt}
\subsection{Low-Power Object Detectors}
\vspace{-5pt}
Nowadays, much attention has been paid to delivering hardware-efficient designs for object detection instead of simply pursuing higher inference quality. To address the design difficulties of real-life applications, a low power object detection challenge in DAC-SDC is proposed to target unmanned aerial vehicle (UAV) applications using embedded platforms, such as NVIDIA TX2 GPU, Ultra96 FPGA, and Xilinx Pynq-Z1 FPGA \cite{xu2019dac}. By examining the winning entries, we notice that all of them adopt one-stage detectors and share similar top-down DNN design approaches in Figure \ref{fig:com_flow}. 
%
As shown in Table \ref{tab:dacsdc_summary}, most of them start from well-established hardware-efficient DNNs, such as ShuffleNet \cite{zhang2018shufflenet}, SqueezeNet \cite{iandola2016squeezenet}, and MobileNet \cite{howard2017mobilenets}, and replace the image classifier with YOLO \cite{redmon2016you,redmon2017yolo9000} or RetinaNet \cite{lin2017focal} back-end for object detection. Other solutions directly adopt the object detection algorithms, such as SSD \cite{liu2016ssd} and YOLO. To deliver hardware-efficient DNNs, they employ input resizing and network pruning to lower the network complexity. Some of the GPU entries use half-precision data format (16-bit) and TensorRT for improved throughput. More aggressive compression is found in FPGA designs because of even tighter resource budgets. DNN parameters are quantized to around 8 bits and some even down to 1 bit. They also cover task partitioning (between host CPU and FPGA), double-pumped DSP (with doubled working frequency in DSP units), tailored pipeline, multithreading, and clock gating to boost hardware performance and energy-efficiency.

\begin{figure*}[t]
\centering
\vspace{-6pt}
\includegraphics[width=1.02\textwidth]{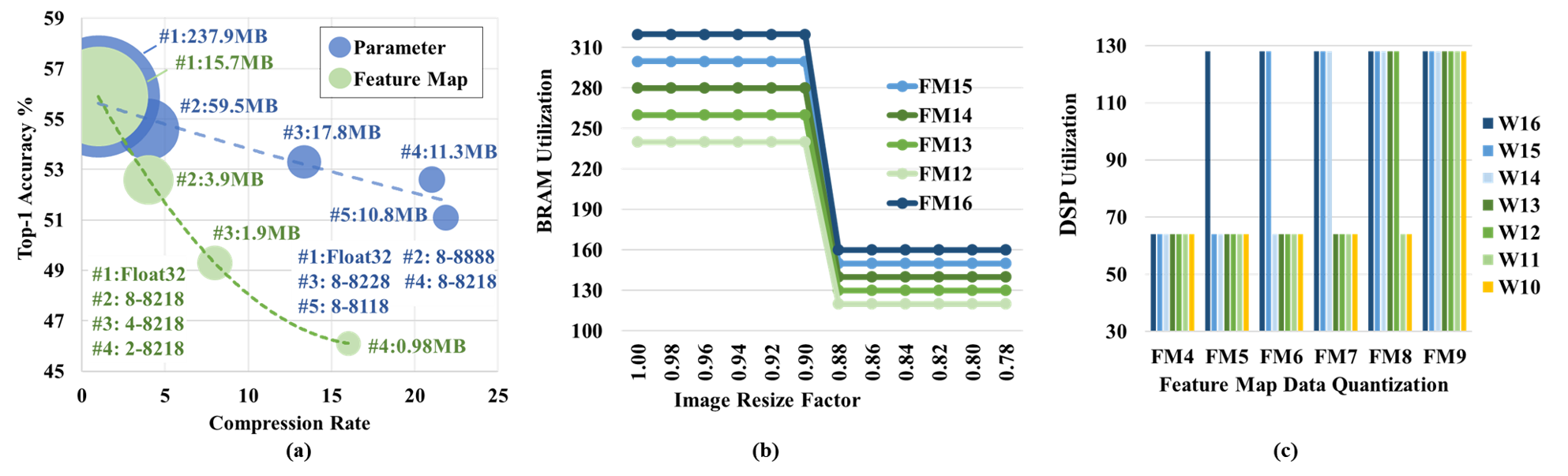}
\vspace{-28pt}
\caption{(a) Accuracy results for the same AlexNet under different compression schemes: blue for parameter compression and green for FM compression. The legend shows the quantization details: 
each quantized model (\#2$\sim$\#5) is denoted as bit precision $p_1$ for FMs across all layers, $p_2$ for $1^{st}$ Conv parameters, $p_3$ for the parameters in $2^{nd}\sim5^{th}$ Convs, $p_4$ for the parameters in $1^{st}\sim2^{nd}$ FCs, and $p_5$ for parameters in the $3^{rd}$ FC in $p_1$-$p_2p_3p_4p_5$ format. (b) BRAM usages of accelerators with the same architecture but 12$\sim$16-bit quantization for feature maps (FM12$\sim$FM16)  and different image resize factors. (c) DSP utilization of accelerator using different quantizations between weights (W) and feature maps (FMs) with the numbers indicating bits allocated.}
\label{fig:motivation}
\vskip -0.2in
\end{figure*}

\vspace{-5pt}
\subsection{Hardware-Aware Neural Network Search}
\vspace{-5pt}
To deliver DNNs for edge devices, there have been growing interests in using neural architecture search (NAS)~\cite{tan2019mnasnet,wu2019fbnet,howard2019searching,cai2018proxylessnas} to automatically find resource constrained DNN targeting edge-platforms. 
To find the efficient networks for a specific platform, \cite{tan2019mnasnet} uses real-time latency by running models on the targeted device instead of latency proxy. Limited by the number of available physical 
{\color{black} devices}, \cite{wu2019fbnet,cai2018proxylessnas} use look-up-table (LUT) to approximate the run-time of models on a specific device. To incorporate human knowledge, \cite{howard2019searching} uses platform-aware NAS to search DNNs for a platform and manually adjust the structure to make it more efficient. Compared to previous hardware-aware NAS methods that target a specific platform, SkyNet can target both embedded GPU and embedded FPGA platforms and capture hardware limitations by using the realistic hardware performance feedback instead of using LUT approximation.

\vspace{-5pt}
\section{Motivations}
\vspace{-5pt}
\label{sec:motivation}
To deliver an even better solution, 
we investigate the potential obstacles in the top-down design flow (Figure \ref{fig:com_flow}) which may hinder further improvements on DNN accuracy and efficiency. We summarize two challenges as follows:
\vspace{-8pt}
\begin{itemize}
\setlength{\itemsep}{-0pt}
    \item [1)] It is difficult to balance the sensitivities of DNN configurations on software and hardware during model compression following the top-down approach. 
    \item [2)] It is difficult to select appropriate reference DNNs at the very beginning of the top-down flow because of the uncertain accuracy variations for a given real-life task.
    \vspace{-10pt}
\end{itemize}
\vspace{-5pt}
The first challenge causes tedious iterative explorations between software and hardware optimizations. With the similar hardware performance (e.g., throughput and latency), DNNs may have different accuracy results as the compression technologies are applied to different network components. We take data quantization as an example. In Figure \ref{fig:motivation} (a), the accuracy results vary significantly between parameter and intermediate feature map (FM) quantization. In this figure, the coordinates of the bubble center represent accuracy and model compression ratio, while the area of a bubble shows data size in megabyte (MB). We scale-up the FM bubble for better graphic effect. By compressing the model from Float32 to fixed point, we reduce 22X parameter size (237.9MB$\rightarrow$10.8MB) and 16X FM size (15.7MB$\rightarrow$0.98MB), respectively. 
In this case, inference accuracy is more sensitive to the FM precision.

\begin{table}[t!]
\vspace{-5pt}
\caption{Accuracy on DAC-SDC dataset using ResNet,
VGG,
and SkyNet backbones and the same back-end for object detection.}
\label{tab:upper-bound}
\begin{center}
\footnotesize
\begin{sc}
\begin{tabular}{l|c|c}
\toprule
Backbone DNN & \# of Parameter & IoU \\
\midrule
ResNet-18  & 11.18M  & 0.61 \\ 
ResNet-34  & 21.28M  & 0.26 \\ 
ResNet-50  & 23.51M  & 0.32 \\ 
VGG-16   & 14.71M  & 0.25 \\ \hline
SkyNet (ours)& \textbf{0.44M} & \textbf{0.73}\\
\bottomrule
\end{tabular}
\end{sc}
\end{center}
\vskip -0.3in
\end{table}

On the other hand, DNNs with similar accuracy may cause differences in hardware. To provide a quantitative analysis, 
Figure \ref{fig:motivation} (b) shows the BRAM (on-chip memory in FPGA) usages with different input sizes and FM quantizations. By reducing the resize factor from 1.00 to 0.78, we can maintain nearly the same DNN accuracy (\textless1.0\% drop), but save half memory when the factor is smaller than 0.9.
Similarly, Figure \ref{fig:motivation} (c) indicates small changes may lead to diverse DSP utilization. By taking the 16-bit FM (FM16) as an example, the required DSPs reduce from 128 to 64 when weights are changed from 15-bit (W15) to 14-bit (W14).

For the second challenge, it is difficult to select a reference DNN with relatively high accuracy upper bound on a given task. The DNNs with impressive accuracy on published datasets (e.g., CIFAR-10/100 and ImageNet) may not be always suitable. We evaluate the accuracy of popular DNNs on DAC-SDC object detection dataset and list the results in Table \ref{tab:upper-bound}. With the same box regression part, these DNNs show no clear 
{\color{black} connection between}
their parameter size and inference accuracy after adequate training.
Thus, it is not easy to select a promising reference model for a given task. 

\begin{figure*}[t]
\centering
\vspace{-5pt}
\includegraphics[width=0.83\textwidth]{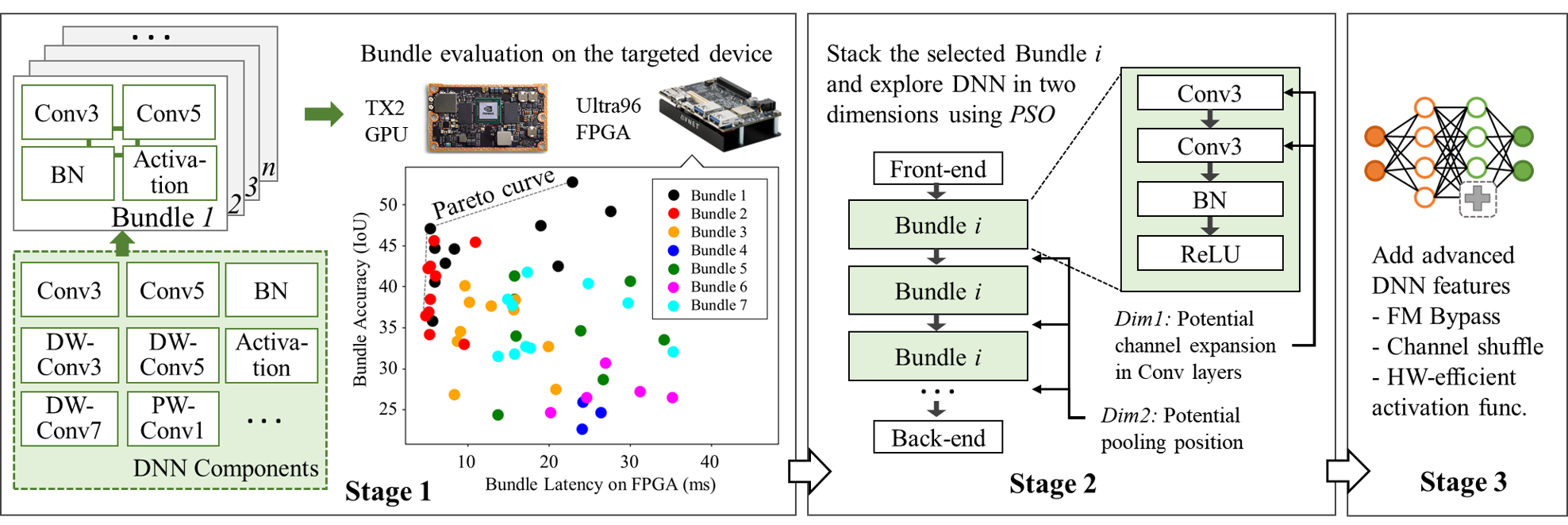}
\vspace{-10pt}
\caption{The proposed bottom-up DNN design flow to deliver hardware-efficient DNNs for embedded systems in three stages.}
\label{fig:bottomup_flow}
\vskip -0.15in
\end{figure*}

\vspace{-5pt}
\section{A Bottom-Up Design Approach}
\vspace{-5pt}
Motivated by the discussed challenges in Section \ref{sec:motivation}, we propose a bottom-up approach to leverage the hardware-efficient DNN design for embedded systems. It is a three-stage approach as shown in Figure \ref{fig:bottomup_flow}.

\subsection{Stage 1: Bundle Selection and Evaluation}
\vspace{-5pt}

This flow starts with building the hardware-aware basic blocks, called Bundles. From a software perspective, a Bundle is a set of sequential DNN layers, which can be repeatedly stacked 
{\color{black} for constructing}
DNNs. 
{\color{black} From the} hardware perspective, a Bundle is a set of IPs to be implemented on hardware.
To capture the hardware constraints, Bundles need to be evaluated on targeted embedded systems
for collecting realistic latency (for both FPGA and GPU) and resource utilization (for FPGA) results.

In the first stage, we enumerate DNN components (such as Conv, pooling, activation layers, etc.) and assemble them into Bundle $1 \sim n$. Each Bundle is then implemented and evaluated in targeted hardware devices for hardware performance metrics. To get their potential accuracy contribution, we build DNN sketches with fixed front- and back-end structures based on given tasks, and respectively insert one type of Bundle (with replications) in the middle. We limit one type of Bundle for one DNN sketch to guarantee its hardware efficiency.
Then, DNN sketches are fast trained using targeted datasets to find out the ones with relatively high accuracy.
%
By targeting the object detection task, for example, we can concatenate a input resizing unit (front-end) and a bounding box regression (back-end) with the selected Bundle to build a DNN sketch. The number of training epochs may vary from different datasets as a 20-epoch-training can distinguish sketches using the DAC-SDC dataset (with 100K images), while 5 epochs are enough if using Cifar-10 dataset. We have also seen similar strategies in \cite{jiang2019accuracy} distinguish candidates by a 25-epoch-training on a subset of ImageNet. 
%
At last, the most promising Bundles located in the Pareto curve are selected for the next stage.

\vspace{-5pt}
\subsection{Stage 2: Hardware-Aware DNN Search}
\vspace{-5pt}
During DNN search, the inputs include
the software and hardware metrics (e.g., DNN accuracy and throughput performance) and the targeted hardware platforms while the outputs are DNN candidates which meet the software and hardware requirements. 
%
%
To solve such a multi-objective optimization problem, we propose a group-based particle swarm optimization (PSO) evolutionary algorithm to discover proper DNN candidates since literature has demonstrated the validity of using evolutionary methods to discover DNNs with state-of-the-art accuracy \cite{real2019regularized,elsken2018efficient}.
From the design methodology perspective, SkyNet can be extended to support other optimization algorithms and meet the needs of different scenarios. 

In the proposed group-based PSO algorithm, each individual DNN is regarded as a particle, and all active DNNs during the search contribute to the swarm.
Since we only use one type of Bundle in 
{\color{black} each} DNN, DNNs composed by the same type of Bundle are considered as a particle group.
In order to maintain evolution stability, a DNN only evolves within its own group.
We label the group optimal position as $P_{group}^{i}$ within the $i$-$th$ group, meaning such DNN has the best fitness value evaluated under given conditions. We denote a DNN particle {\color{black} $j$} within group $i$ as $n_j^i$ and each $n_j^i$ has a pair of feature vectors $(fv1,fv2)$ to illustrate two hyper-parameters regarding DNN structure.
$fv1$ represents the number of channels of each Bundle replication; and $fv2$ describes the pooling position between Bundles. Both feature vectors with dimension equal to the number of stacked Bundle{\color{black}s} in $n_j^i$, and both of them affect accuracy and hardware performance. 
To locate the best DNN candidates, we propose Algorithm 1 with the following major components:

\textit{Population generation}. An initial network population $P$ (a set of DNN candidates) is generated with $\mathcal{M}$ groups and $\mathcal{N}$ networks for each group.
The search contains $\mathcal{I}$ iterations and in the $itr$-th iteration, all networks are fast trained for $e_{itr}$ epochs, where $e_{itr}$ increases with $itr$.

\textit{Latency estimation}.
We perform a platform-specific latency estimation.
For GPUs, we directly measure the inference latency on the training GPU, and scale latency to the targeted GPU for deployment if the target GPU is different from the training one.
For FPGAs, we follow a predefined IP-based DNN accelerator template \cite{hao2019fpga} for hardware performance evaluation. Layer-specific IPs are implemented in hardware and shared by corresponding DNN layers.
To maximize the performance, IPs are configured to fully consume the available resources. We then collect the end-to-end performance and resource overhead of each  DNN from an FPGA high level synthesis tool.

\textit{Fitness value}.
After network training and latency estimation, we calculate the fitness value for each network $n_j^i$ as:
\vspace{-4pt}
\begin{equation}
\footnotesize
\vspace{-2pt}
    Fit_j^i = Acc_j^i + \alpha \cdot ( Est(n_j^i) - Tar  )
\vspace{-2pt}
\end{equation}
where $Acc_j^i$ is the validation accuracy of $n_j^i$ and 
$Est(n^i_j)$ represents the latency on hardware; $Tar$ is the targeted latency. Parameters $\alpha$ ($\alpha < 0$) is used to balance between network accuracy and hardware performance.

\textit{Velocity calculation and particle update}.
In standard PSO, the updated velocity of a particle is calculated every iteration based on the current velocity, the velocities toward the local and the global best positions. Particles can move to a better position 
{\color{black} with assigned}
 probabilities following the updated velocity.
Similarly, in our case, DNNs in the same group update their positions (meaning network structures represented by feature vectors) based on the current design, the local best design (the best one across all passing iterations), and the group best design.   
To determine the velocity toward the local best $V_{local}$ and the group best $V_{group}$, we compute the differences between positions of current and the local/group best designs. Since each position is represented by $(fv1, fv2)$, position differences can be captured by the mismatch of layer expansion factors $fv1$ and pooling spots $fv2$, respectively. Then, with the velocities known, we start evolving the current network by updating its position toward the local and the group best by a random percentage. 

\vspace{-5pt}
\subsection{Stage 3: Feature Addition}
\vspace{-5pt}
More advanced DNN design features are added if hardware metrics allow. For example, we can include a bypass from low-level features to high-level features along with feature map reordering~\cite{redmon2017yolo9000} to improve small object detection. We can also replace ReLU with ReLU6~\cite{sandler2018mobilenetv2} to enhance hardware efficiency. More discussions are provided in the next section.

\setlength{\textfloatsep}{10pt}
\begin{algorithm}[t!]
\renewcommand\baselinestretch{0.8}\selectfont
\footnotesize
$P \leftarrow$ InitialPopulation($\mathcal{M}$, $\mathcal{N}$)\\
\While{$ itr < \mathcal{I}$}{
\textbf{FastTraining}($P$, $e_{itr}$)\\
$Fit_j^i \leftarrow$ \textbf{GetFitnessVal}($P$) //evaluate all candidates\\ 
\For{each group $i$}{
\textbf{GroupRank}($i$) //rank candidates in group i\\
$N_{group}^i \leftarrow$ \textbf{GroupBest}($i$) //select the best one in group i\\
//get the group best position\\
 $P_{group}^i(fv1, fv2) \leftarrow$\textbf{GetPosition}($N_{group}^i$) \\
 \For{each candidate $n_j^i(itr)$ in group $i$} {
    //rank $n^i_j$ across all passing iterations\\
    \textbf{LocalRank}($i,j$) \\
    $N_{local}^{ij} \leftarrow$ \textbf{LocalBest}($i,j$) \\
    //get the local best position\\
    $P_{local}^{ij}(fv1, fv2) \leftarrow$\textbf{GetPosition}  ($N_{local}^{ij}$) \\
    //get the current position\\
    $P^{i}_j(fv1, fv2) \leftarrow$\textbf{GetPosition}  ($n_{j}^{i}(itr)$) \\
    //get the velocity toward the local and the group best\\
    $V_{local} \leftarrow$\textbf{GetV}($P^{i}_j,P_{local}^{ij}$) \\
    $V_{group} \leftarrow$\textbf{GetV}($P^{i}_j,P_{group}^{i}$) \\
    $n_j^i(itr+1) \leftarrow$ \textbf{Evolve}($n_j^i(itr)$, $V_{local}$, $V_{group}$) \\
}}
}
\caption{The bottom-up DNN design with PSO}
\label{algorithm1}
\end{algorithm}

\vspace{-5pt}
\section{SkyNet}
\vspace{-5pt}
\begin{figure*}[t!]
\centering
\vspace{-10pt}
\includegraphics[width=0.88\textwidth]{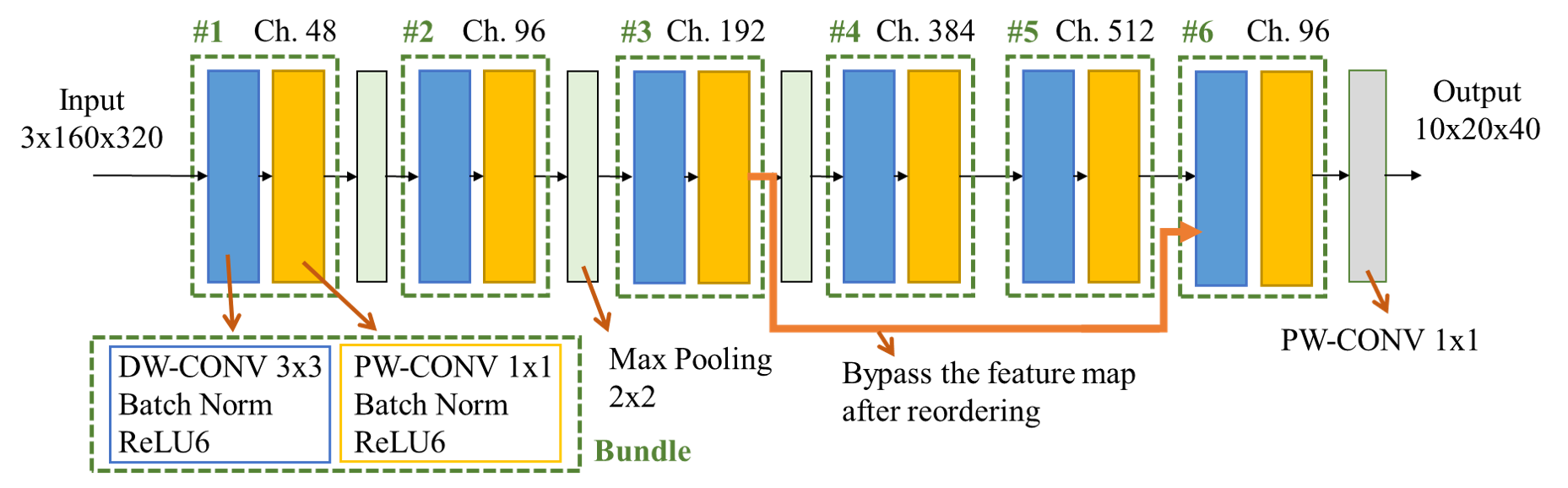}
\vspace{-10pt}
\caption{SkyNet backbone (model C in Table \ref{tab:skynet_arch}) generated by stacking six of the selected Bundle (circled by green dashed line) with DNN components as: DW-Conv3, PW-Conv1, BN, and ReLU6. The number of output channels is listed on top of each Bundle denoted as Ch. Three 2$\times$2 pooling layers are inserted. The bypass is highlighted in orange, which passes feature maps generated by the Bundle \#3 directly to the last Bundle. The feature map reordering is also performed along with the bypass.}
\vspace{-10pt}
\label{fig:SkyNet}
\end{figure*}


\subsection{SkyNet Architecture for object detection}
\vspace{-5pt}
Following the proposed flow, the best Bundle is selected as a combination of 3$\times$3 depth-wise Conv layer (DW-Conv3~\cite{howard2017mobilenets}), 1$\times$1 point-wise Conv layer (PW-Conv1), batch normalization layer (BN~\cite{ioffe2015batch}), and ReLU6. By repeatedly stacking this Bundle, we generate three backbones shown in Table \ref{tab:skynet_arch} for object detection in DAC-SDC. These networks share the same chain structure but with different configurations of feature map bypass. For model A, no bypass is included; while for the model B and C, output feature maps of Bundle \#3 are fed into the Bundle \#6. SkyNet also adapts the YOLO detector head by removing the classification output and use two anchors for bounding box regression.

\vspace{-5pt}
\subsection{Feature Map Bypass, Reordering, and ReLU6}
\begin{figure}[t!]
\centering
\includegraphics[width=0.35\textwidth]{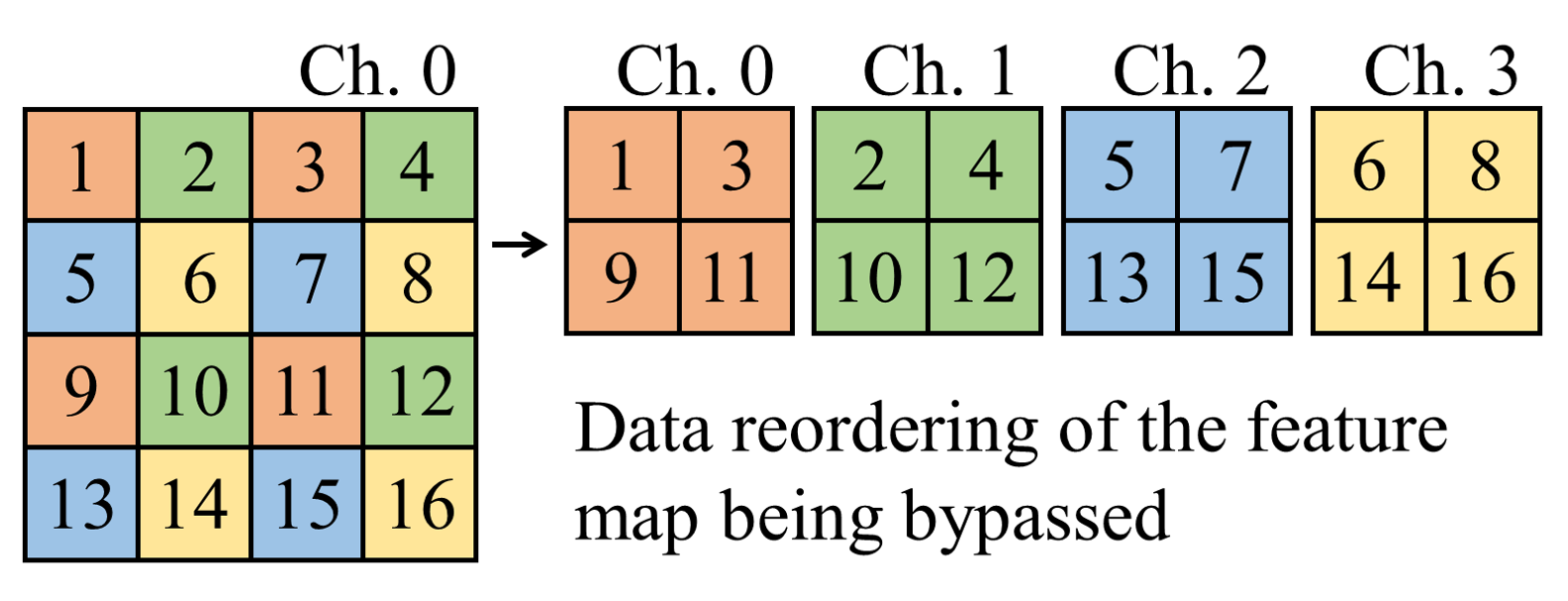}
\vspace{-10pt}
\caption{Feature map reordering from $1\times4\times4$ to $4\times2\times2$ with shrunken width and height but expanded number of channels. There is no information loss compared to pooling operation. In addition, this reorder pattern also ensures larger receptive field.}
\label{fig:reorder}
\end{figure}

\begin{figure}[th!]
\centering
\includegraphics[width=0.43\textwidth]{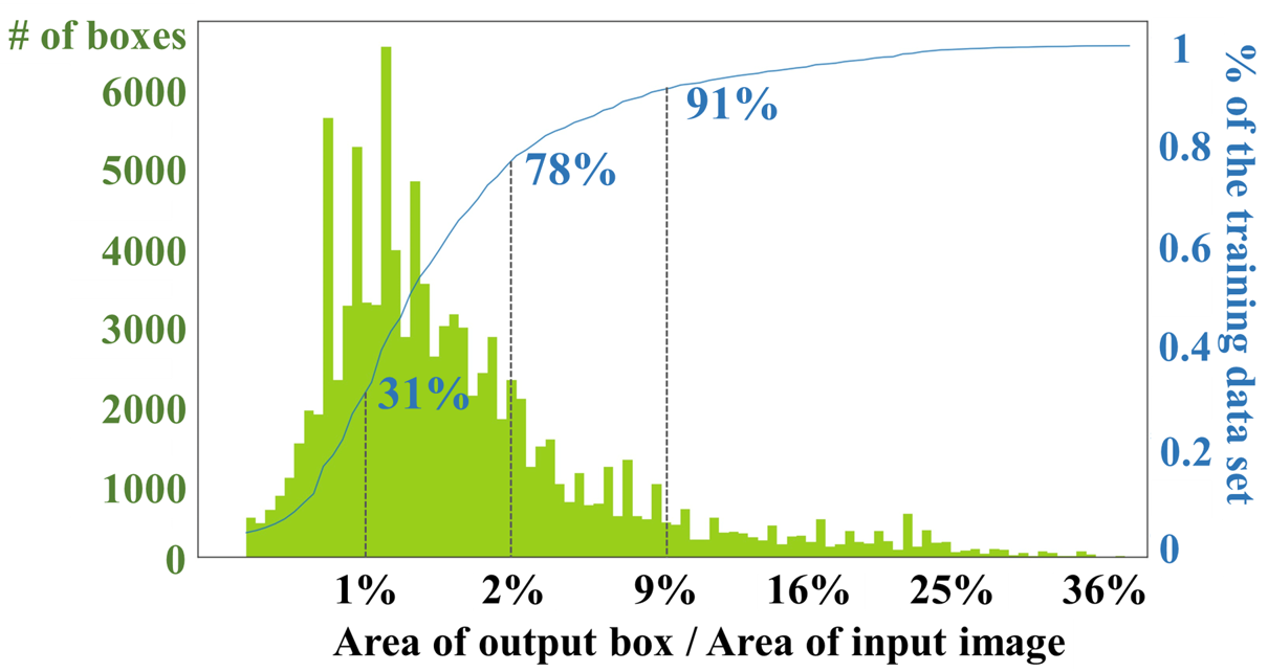}
\vspace{-10pt}
\caption{The distribution of bounding box relative size in DAC-SDC training dataset. We capture the bounding box relative size by computing the ratio of output bounding box size divided by the input image size. The green bars show the ratio distribution, and the blue curve shows the corresponding cumulative distribution.}
\label{fig:data_analysis}
\end{figure}

\vspace{-5pt}
By examining the DAC-SDC training data, we keep a record of the size ratio between the output bounding box and the input image and present a distribution diagram in Figure \ref{fig:data_analysis}. It clearly shows that 91\% of the objects to be detected are less than 9\% of the original input image size, and 31\% of them are even smaller than 1\% of the input image size. It means the majority of objects inside this dataset are small objects and we need to provide additional DNN features accordingly.
So, we add feature map bypass and reordering to enhance the ability of detecting small object (model B and C). The bypass helps to keep small object features in the later part (closer to the output layer) of the DNN by adding low-level high-resolution feature maps. Also, it is beneficial to have multiple feature maps (from different layers) before generating the bounding boxes. Since the bypass crosses a pooling layer (highlighted in Figure \ref{fig:SkyNet}), we use reordering (shown in Figure \ref{fig:reorder}) to align the size of original feature map (generated by the Bundle \#5) and the low-level feature without losing information. 
%
The other feature to improve hardware efficiency is the ReLU6, which clips output range to $[0,6]$. Since ReLU6 generates much smaller data range compared to the original ReLU ($[0,+ \infty)$), less bits are required to represent intermediate FMs. It also helps to better implement lower-precision floating point in embedded GPUs and fixed-point data format in embedded FPGAs.

\begin{table}[t]
\vspace{-10pt}
\caption{The SkyNet architecture with number of channels shown in the bracket. Each convolutional layer except the last one is followed by a BN and a ReLU (omitted for conciseness).
}
\begin{small}
\begin{sc}
\footnotesize
\newcommand{\tabincell}[2]{\begin{tabular}{@{}#1@{}}#2\end{tabular}}
\label{tab:skynet_arch}
\begin{center}
\begin{tabular}{c|c|c|c}
\toprule
\multicolumn{4}{c}{Configurations of SkyNet} \\ 
\hline
A   & B   & C & Bundle \\ \hline
\multicolumn{3}{c}{input (3$\times$160$\times$360 color image)} \\ \hline
\multicolumn{3}{c|}{\begin{tabular}[c]{@{}c@{}}DW-Conv3 (3) \\ PW-Conv1 (48)\end{tabular}} & \#1 \\ \hline
\multicolumn{3}{c}{2$\times$2 max-pooling} \\ \hline
\multicolumn{3}{c|}{\begin{tabular}[c]{@{}c@{}}DW-Conv3 (48) \\ PW-Conv1 (96)\end{tabular}} & \#2\\ \hline
\multicolumn{3}{c}{2$\times$2 max-pooling} \\ \hline
\multicolumn{3}{c|}{\begin{tabular}[c]{@{}c@{}}DW-Conv3 (96) \\ PW-Conv1 (192) \\ \textit{\textbf{$[$Bypass Start$]$ FM Reordering}} (768) \end{tabular}} & \#3\\ \hline
\multicolumn{3}{c}{2$\times$2 max-pooling} \\ \hline
\multicolumn{3}{c|}{\begin{tabular}[c]{@{}c@{}}DW-Conv3 (192) \\ PW-Conv1 (384) \\ \end{tabular}} & \#4 \\ \hline
\multicolumn{3}{c|}{\begin{tabular}[c]{@{}c@{}}DW-Conv3 (384) \\ PW-Conv1 (512) \\ \end{tabular}} & \#5\\ \hline

\begin{tabular}[c]{@{}c@{}}
\tabincell{c}{  \\ $\qquad $  } \\ \end{tabular} & \begin{tabular}[c]{@{}c@{}} \textit{\textbf{$[$Bypass End$]$}} \\ \textit{\textbf{FM Concatenated}} \\ \tabincell{c}{DW-Conv3 \\ (512+768)}  \\ PW-Conv1 (48) \\ \end{tabular} & \begin{tabular}[c]{@{}c@{}} \textit{\textbf{$[$Bypass End$]$}} \\ \textit{\textbf{FM Concatenated}}\\ \tabincell{c}{DW-Conv3 \\ (512+768)} \\ PW-Conv1 (96) \\
\end{tabular} & \#6\\ \hline
\multicolumn{4}{c}{\begin{tabular}[c]{@{}c@{}}PW-Conv1 (10) \\ Back-end for bounding box regression\end{tabular}}
\\ 
\bottomrule
\end{tabular}
\end{center}
\end{sc}
\end{small}
\vskip -0.1in
\end{table}

\begin{figure*}[t!]
\centering
\includegraphics[width=0.9\textwidth]{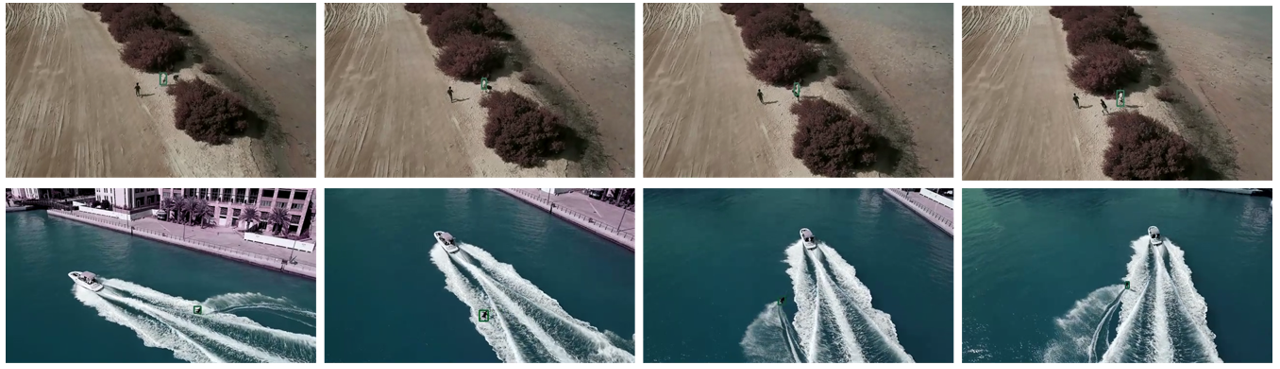}
\vspace{-10pt}
\caption{Object detection results generated by SkyNet on DAC-SDC dataset. Challenges include to detect small objects and distinguish multiple similar objects (e.g., images in the first row).}
\label{fig:dataset}
\vskip -0.3in
\end{figure*}

\vspace{-5pt}
\section{Experiments on DAC-SDC}
\vspace{-5pt}
DAC-SDC features a single object detection challenge for embedded systems, which include embedded GPUs (NVIDIA TX2) and FPGAs (Pynq-Z1 and Ultra96) with very low energy consumption. The goal is to consider the most appropriate needs of UAV applications, such as capability of real-time processing, energy efficiency, and detection accuracy. To better reflect real-life challenges, images of the dataset are captured by UAVs in the real environment. The whole dataset is divided by two parts: the training dataset with 100,000 images with objects of interest across 12 main categories and 95 sub-categories, and the hidden test set for official evaluation with 50,000 images that only the contest organizers could access \cite{DAC-SDC-dataset}. Results generated by SkyNet are shown in Figure \ref{fig:dataset}.
In DAC-SDC'19, 52 GPU teams and 58 FPGA teams participated worldwide creating a very intense competition. Our SkyNet design has successfully delivered the best inference accuracy and total score for both GPU and FPGA tracks.

\subsection{Ablation Study}
\vspace{-5pt}
We perform an ablation study on DAC-SDC dataset to analyze these three configurations of SkyNet (Model A, B, and C listed in Table \ref{tab:skynet_arch}). By combining two activation functions (ReLU and ReLU6), six configurations of SkyNet are evaluated. We train these models in an end-to-end fashion using multi-scale training with the learning rate starting from 1e-4 to 1e-7. We apply stochastic gradient descent (SGD) to update parameters. To further enrich the training data, we use data augmentations to distort, jitter, crop, and resize inputs with size 160$\times$320. The accuracy results are presented in Table \ref{tab:training_accuracy}, where SkyNet C - ReLU6 reaches the highest IoU (0.741) on the validation set. Therefore, we use this model as the proposed design for the following experiments.

\begin{table}[t]
\vspace{-10pt}
\caption{Validation accuracy of SkyNet.}
\label{tab:training_accuracy}
\begin{center}
\begin{sc}
\begin{small}
\newcommand{\tabincell}[2]{\begin{tabular}{@{}#1@{}}#2\end{tabular}}
\begin{tabular}{l|c|c}
\toprule
DNN Model & Parameter Size& IoU \\
\midrule
\tabincell{c}{SkyNet A - ReLU} & 1.27 MB & \tabincell{c}{0.653} \\ 
\tabincell{c}{SkyNet A - ReLU6} & 1.27 MB & \tabincell{c}{0.673} \\
\tabincell{c}{SkyNet B - ReLU} & 1.57 MB &\tabincell{c}{0.685} \\ 
\tabincell{c}{SkyNet B - ReLU6} & 1.57 MB &\tabincell{c}{0.703} \\ 
\tabincell{c}{SkyNet C - ReLU} & 1.82 MB &\tabincell{c}{0.713} \\ 
\tabincell{c}{SkyNet C - ReLU6} & 1.82 MB &\tabincell{c}{\textbf{0.741}} \\ 
\bottomrule
\end{tabular}
\end{small}
\end{sc}
\end{center}
\vskip -0.1in
\end{table}

\vspace{-5pt}
\subsection{Evaluation Criteria}
\vspace{-5pt}
Comprehensive evaluations are introduced in DAC-SDC, covering detection accuracy (IoU), throughput (FPS), and energy consumption. To identify the best design, a total score is calculated following Equation \ref{eq:iou} to \ref{eq:TS}.
Assuming there are $I$ registered teams and $K$ images in the test set, the IoU score for team $i$, denoted as $R_{IoU_i}$, is calculated as: 
\begin{equation}
\label{eq:iou}
\footnotesize
\vspace{-4pt}
    R_{IoU_i} = \frac{ \sum\limits_{k=1}^{K} IoU_{i,k}}{K}
\vspace{-2pt}
\end{equation}

For energy, $\bar{E}_{I}$ is denoted as the average energy consumption of all $I$ entries when performing DNN inference on the test dataset (Equation \ref{eq:E_avg}). The energy score of team $i$ ($ES_{i}$) is then computed using Equation \ref{eq:ES} relating to the ratio between average energy and the energy consumed by this team. $x$ is set to 2 and 10 for FPGA track and GPU track, respectively. 
Eventually, the total score, denoted as $TS_{i}$, is calculated in Equation \ref{eq:TS} including both inference accuracy ($R_{IoU_i}$) and energy consumption ($ES_i$). 

\begin{equation}
\label{eq:E_avg}
\footnotesize
\vspace{-14pt}
    \bar{E}_{I} = \frac{ \sum\limits_{i=1}^{I} E_{i}}{I}
\vspace{-2pt}    
\end{equation}

\begin{equation}
\label{eq:ES}
\footnotesize
\vspace{-8pt}
    ES_{i} = max\{0, 1+0.2\times log_x \frac{\bar{E}_I}{E_i}  \}
\vspace{-0pt}
\end{equation}

\begin{equation}
\label{eq:TS}
\footnotesize
\setlength{\abovedisplayskip}{-10pt}
\setlength{\belowdisplayskip}{-10pt}
\vspace{-4pt}
    TS_{i} = R_{IoU_i} \times (1+ ES_i )
\vspace{-4pt}
\end{equation}

\begin{table*}[t!]
\vspace{-15pt}
\caption{GPU final results from DAC-SDC'19 and '18 using the hidden test set with 50K images, evaluated by a TX2 GPU.}
\vspace{-2pt}
\label{tab:rst_comp_gpu}
\begin{center}
\begin{small}
\begin{sc}
\newcommand{\tabincell}[2]{\begin{tabular}{@{}#1@{}}#2\end{tabular}}
\begin{tabular}{l|c|c|c|c}
\toprule
\textbf{Team Name} &  \textbf{IoU} & \textbf{FPS} & \textbf{Power(W)} & \textbf{Total Score} \\
\midrule
\multicolumn{5}{c}{Results from 2019} \\
\midrule
\tabincell{c}{\textbf{SkyNet} (ours) } & \textbf{0.731} & \textbf{67.33} & 13.50 & \textbf{1.504} \\ 
\hline
\tabincell{c}{Thinker  \cite{thinker}}  & 0.713 & 28.79 & \textbf{8.55} & 1.442 \\ 
\hline
\tabincell{c}{DeepZS  \cite{DeepZS}} &  0.723 & 26.37 & 15.12& 1.422 \\ 
\midrule
\multicolumn{5}{c}{Results from 2018} \\
\midrule
\tabincell{c}{ICT-CAS  \cite{ICT-CAS}}  & 0.698 & 24.55 & 12.58& 1.373 \\ 
\hline
\tabincell{c}{DeepZ  \cite{DeepZ}} & 0.691 & 25.30 & 13.27 & 1.359 \\ 
\hline
\tabincell{c}{SDU-legend  \cite{SDU-Legend}} & 0.685 & 23.64 & 10.31& 1.358 \\ 
\bottomrule
\end{tabular}
\end{sc}
\end{small}
\end{center}
\vskip -0.2in
\end{table*}

\begin{table*}[t!]
\vspace{-5pt}
\caption{FPGA final results in DAC-SDC'19 and '18 using the hidden test set with 50K images. Designs in 2019 are evaluated on a Ultra96 FPGA while designs in 2018 use a Pynq-Z1 FPGA.}
\vspace{-2pt}
\label{tab:rst_comp_fpga}
\begin{center}
\begin{small}
\begin{sc}
\newcommand{\tabincell}[2]{\begin{tabular}{@{}#1@{}}#2\end{tabular}}
\begin{tabular}{l|c|c|c|c}
\toprule
\textbf{Team Name} & \textbf{IoU} & \textbf{FPS} & \textbf{Power (W)} & \textbf{Total Score} \\
\midrule
\multicolumn{5}{c}{Results in 2019} \\
\midrule
\tabincell{c}{\textbf{SkyNet} (ours) \\ }  & \textbf{0.716} & 25.05 & 7.26 & \textbf{1.526} \\ 
\hline
\tabincell{c}{XJTU\_Tripler \cite{XJTU}}  & 0.615 & 50.91 & 9.25 & 1.394 \\ 
\hline
\tabincell{c}{SystemsETHZ \cite{systemsETHZ19}}  & 0.553 & \textbf{55.13} & 6.69 & 1.318 \\ 
\midrule
\multicolumn{5}{c}{Results in 2018} \\
\midrule
\tabincell{c}{TGIIF \cite{TGIIF}} & 0.624 & 11.96 & 4.20& 1.267 \\ 
\hline
\tabincell{c}{SystemsETHZ \cite{systemsETHZ}}  & 0.492 & 25.97 & \textbf{2.45} & 1.179 \\ 
\hline
\tabincell{c}{iSmart2 \cite{iSmart2}}  & 0.573 & 7.35 &2.59& 1.164 \\ 
\bottomrule
\end{tabular}
\end{sc}
\end{small}
\end{center}
\vskip -0.0in
\end{table*}

\subsection{GPU Implementation}
\vspace{-5pt}
\label{sec:gpu_impl}
For the TX2 GPU implementation, we keep all network parameters using Float32 to maintain the best inference accuracy. Since most of the compute-intensive parts of DNN inference are handled by NVIDIA cuDNN, which leaves little space for customized improvement, we start optimizing our design on a system-level. 

The whole procedure of running SkyNet contains four steps as: 1) input fetching from the flash storage in a unit of batch; 2) image pre-process which includes input resizing and normalization; 3) DNN inference; and 4) post-process to generate bounding boxes and buffer results in DDR memory. The most straightforward way is to execute these steps in serial but with the cost of low resource utilization and poor throughput performance. In our design, we first merge step 1 and 2 in pre-process and enable multithreading technology to execute these steps in a pipelined fashion as shown in Figure~\ref{fig:task_partition}. 
%
We use NVIDIA System Profiler (L4T) to capture the latency results. In average, the proposed system-level optimizations enable a 3.35X speedup compared to the original serial design and help our design reach the highest throughput performance, peaking at 67.33 FPS. 

\vspace{-5pt}
\subsection{FPGA Implementation}
\vspace{-5pt}
\label{sec:fpga_impl}
To implement DNNs on FPGA, we suffer even scarcer resource budgets, as the theoretical peak performance provided by Ultra96 FPGA (144 GOPS @200MHz) is much lower than the TX2 GPU (665 GFLOPS @1300MHz). By using the proposed bottom-up design flow, hardware limitations have already captured by the Bundle design and the Bundle is instantiated on FPGA as a single customized hardware IP. Since the proposed network is structured by the same type of Bundle, this IP can be shared across different layers to cope with the resource constraints.
Still, we need more optimizations to further enhance the performance. 

\vspace{-5pt}
\subsubsection{Quantization, Batch Process, and Tiling}
\vspace{-5pt}
\begin{table}[t]
\footnotesize
\newcommand{\tabincell}[2]{\begin{tabular}{@{}#1@{}}#2\end{tabular}}
\centering
\caption{Validation accuracy results regarding different quantization schemes during FPGA implementation} \label{tab:quantization}
\begin{tabular}{c|c|c|c}
\toprule
 Scheme & Feature Map & Weight  & Accuracy (IoU)  \\ \midrule
0 & Float32 & Float32 & 0.741 \\
1 & 9 bits & 11 bits & 0.727 \\
2 & 9 bits & 10 bits & 0.714 \\
3 & 8 bits & 11 bits & 0.690 \\
4 & 8 bits & 10 bits & 0.680 \\
\bottomrule
\end{tabular}
\end{table}

Since fixed-point representation is more favorable in FPGA design, we quantize the FMs and weights from Float32 to fixed point and explore different quantization schemes in Table \ref{tab:quantization}. 
After quantization, the SkyNet backbone suffers different levels of accuracy drop from 1.4\% to 6.1\% in scheme 1 to 4. We finally pick scheme 1 as accuracy has higher weight in the total score calculation (Equation \ref{eq:TS}).

Since network parameters can not be accommodated by the FPGA on-chip memory (which is BRAM with only 0.95 MB available), we have to store them in the external memory (DRAM), which easily makes the memory access bandwidth a bottleneck. To mitigate the bandwidth demand, input batch process is applied to exploit data reuse opportunities, where a certain number (which equals to the batch size) of input images are assembled before sending to FPGA for DNN inference, so that task size (the number of images being processed at one time) increases while consuming the same amount of network parameters from DRAM. 

\begin{figure}[t]
\centering
\includegraphics[width=0.45\textwidth]{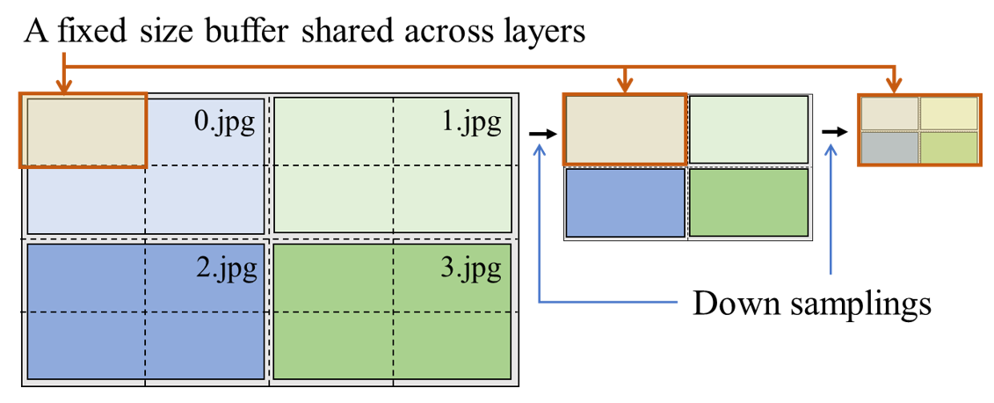}
\vspace{-10pt}
\caption{The proposed batch and tiling design to increase the data reuse opportunity and avoid on-chip memory waste.}
\label{fig:tiling_batch}
\vspace{-5pt}
\end{figure}

With larger batch size, the process of network inference asks for larger amount of FPGA on-chip memory to buffer intermediate FMs. Since our implementation is based on an IP-shared structure, buffers instantiated on FPGA are shared by different layers, which means the buffer may not be large enough for the FMs generated by the first few layers while too large for the last few layers as FMs get smaller after pooling. 
To solve this problem, we propose an input tiling and batch scheme as shown in Figure \ref{fig:tiling_batch}.
Four inputs are stitched to form a larger input which can be processed as an entirety. With the tiling and batch process, it is possible to use one shared buffer across different layers without changing its size. The proposed solution inherits the benefit from batch process to allow better reuse of DNN weights and it eliminates the possible waste of unused buffer space.

\begin{figure*}[ht!]
\centering

\includegraphics[width=0.9\textwidth]{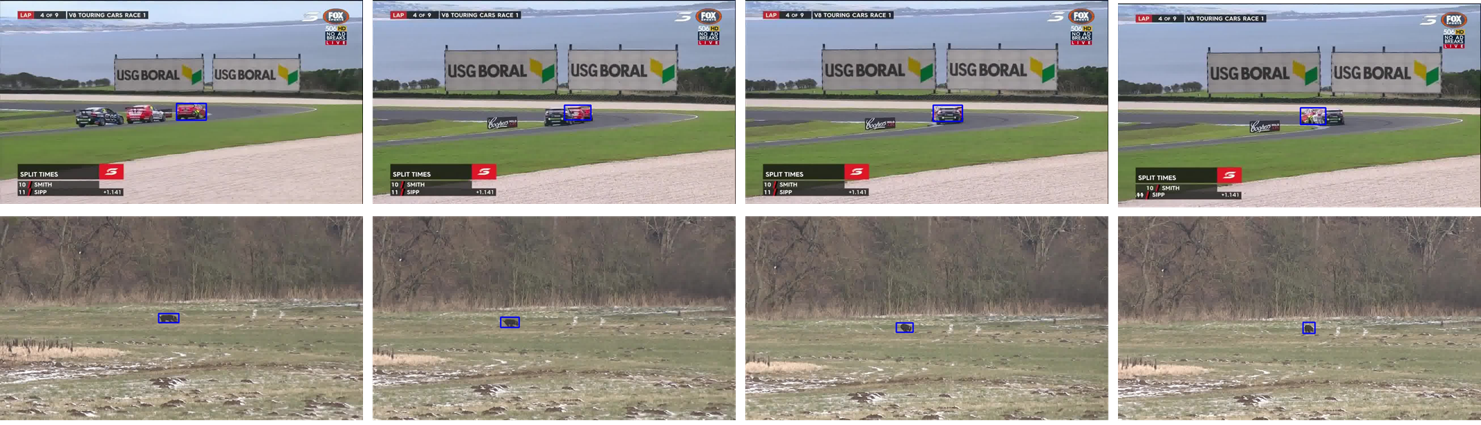}
\vspace{-10pt}
\caption{Object tracking results generated by SkyNet on GOT-10K dataset.}
\label{fig:dataset_got10 }
\vspace{-15pt}
\end{figure*}

\begin{figure}[t]
\centering
\includegraphics[width=0.45\textwidth]{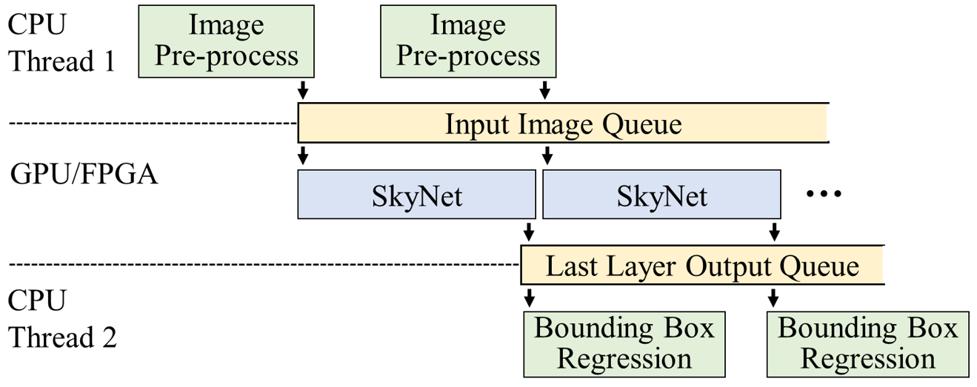}
\vspace{-10pt}
\caption{Task partitioning in SkyNet implementation on TX2 GPU and Ultra96 FPGA.}
\label{fig:task_partition}
\end{figure}

\vspace{-5pt}
\subsubsection{Layer Fusion, Memory Hierarchy, and Task Partitioning}
\vspace{-5pt}
To avoid dealing with the floating-point operations (e.g., inverse-square root) in BN layer, we use layer fusion to merge both parameters from Conv and its successive BN offline. So, there are no separated BN layers nor expensive floating-point operations required during DNN inference.

With hardware resources shared by DNN layers, the intermediate results need to be swapped in/out between on-chip and external memory. To boost the performance, we instantiate the selected Bundle on hardware and implement a five-stage pipeline with Load, EXE\_CONV3, EXE\_CONV1, EXE\_Pooling, and WriteBack stages. By using Ping-pong buffers between memory and computation units, data transfer (in Load and WriteBack stages) can be fully overlapped by computation latency. Regarding the data transfer between adjacent execution stages (with ``EXE'' prefix), we keep data on-chip without going through external memory.

To fully utilize the available computational resource, we also implement task partitioning on the Ultra96. The whole design is shown in Figure \ref{fig:task_partition}, which is highly similar to our GPU design. Workloads are distributed to both CPU and FPGA and creating a system-level pipeline. With all three tasks (pre-process, SkyNet inference, and post-process)  overlapped, our FPGA design can reach 25.05 FPS.

\vspace{-5pt}
\subsection{Result Comparison}
\vspace{-5pt}
After implementing the proposed DNN on GPU and FPGA following the strategies mentioned in Section \ref{sec:gpu_impl} and \ref{sec:fpga_impl}, our designs are evaluated by the DAC-SDC organizers using the hidden test set. As shown in Table \ref{tab:rst_comp_gpu} and \ref{tab:rst_comp_fpga}, we present the comparison results with the top-3 teams in DAC-SDC'19 and '18. In our GPU design, SkyNet outperforms all other competitors by delivering the best accuracy (0.731), throughput performance (67.33), and total score (1.504). In terms of the FPGA design, SkyNet also reaches the best accuracy and gets the highest total score.

\vspace{-5pt}
\section{SkyNet Extension on GOT-10K}
\vspace{-5pt}
Since SkyNet can deliver real-time object detection on embedded systems, we setup experiments on the GOT-10k benchmark \cite{huang2018got} to demonstrate its potential on object tracking. 
GOT-10k is a large high-diversity database for generic object tracking with rich motion trajectory and wide coverage of object classes. 
Models are evaluated with two metrics in GOT-10k as average overlap (AO) and success rate (SR). AO is defined as the mean of IoU between prediction and ground truth bounding boxes, while SR is defined as the proportion of predictions where the IoU is beyond some threshold. 
During evaluation, Got-10K only provides the ground truth bounding box in the first frame and expect trackers to keep tracking on the same object for subsequent frames by predicting bounding boxes.
The predictions will then be evaluated by the Got-10K server. In this section, we integrate the SkyNet backbone with two of the state-of-the-art trackers (SiamRPN++ and SiamMask) and evaluate its capability of real-time tracking.

\vspace{-5pt}
\subsection{Evaluation Using SiamRPN++}
\vspace{-5pt}
\label{sec:siamRPN++}
Siamese network is one of the most popular network structures for building object trackers. The Siamese trackers locate the object by the correlation between features extracted from the exemplar image and search image, where DNN-based feature extraction plays an important role. SiamRPN++ \cite{li2018siamrpn++} is the first Siamese tracker that has been proven to profit from using DNN backbones with different capacities as long as they are properly trained. To evaluate the performance of different backbones, we train three SiamRPN++ trackers with AlexNet, ResNet-50, and SkyNet backbones on GOT-10k. We maintain the size of exemplar and search images as 127$\times$127 and 255$\times$255 (128$\times$128 and 256$\times$256 for SkyNet for better implementation efficiency), respectively, and we set the learning rates start from 1e-3 to 1e-5. Results are shown in Table \ref{tab:siamrpngot10k} where SkyNet achieves nearly the same quality (AO and SR) as the ResNet-50 backbone but much better speed (1.59X faster).

\begin{table}[t]
\footnotesize
\centering
\vspace{-10pt}
\caption{Performance of SiamRPN++ trackers on GOT-10k with different backbones evaluated on single NVIDIA 1080Ti.} 
\vspace{5pt}
\label{tab:siamrpngot10k}
\begin{tabular}{c|c|c|c|c}
\toprule
 Backbone & $AO$ & $SR_{0.50}$ & $SR_{0.75}$ & $FPS$ \\ \midrule
AlexNet & 0.354 & 0.385 & 0.101 & 52.36\\
ResNet-50 & 0.365 & 0.411 & 0.115 & 25.90\\
SkyNet & 0.364 & 0.391 & 0.116 & 41.22\\
\bottomrule

\end{tabular}
\vskip -0.2in
\end{table}

\vspace{-5pt}
\subsection{Evaluation Using SiamMask}
\vspace{-5pt}
SiamMask \cite{wang2019fast} is another Siamese tracker which outperforms SiamRPN++ by incorporating image segmentation for object tracking tasks. Since information of the segmentation is not provided, it cannot be directly trained with GOT-10k dataset. Instead, we perform training using Youtube-VOS dataset \cite{xu2018youtube} and apply object tracking on Got-10K to compare the performance of different backbones using the SiamMask structure. We maintain the same input size setup as Section \ref{sec:siamRPN++} and apply the learning rates from 1e-3 to 1e-4. As shown in Table \ref{tab:siammaskgot10k}, the proposed SkyNet backbone outperforms ResNet-50 in all metrics when using SiamMask tracker with better tracking quality and 1.73X speedup.

\begin{table}[t]
\footnotesize
\newcommand{\tabincell}[2]{\begin{tabular}{@{}#1@{}}#2\end{tabular}}
\centering
\caption{Performance of SiamMask trackers on GOT-10k with different backbones evaluated on single NVIDIA 1080Ti.} \label{tab:siammaskgot10k}
\vspace{5pt}
\begin{tabular}{c|c|c|c|c}
\toprule
Backbone & $AO$ & $SR_{0.50}$ & $SR_{0.75}$ & $FPS$ \\ \midrule
ResNet-50 & 0.380 & 0.439 & 0.153 & 17.44\\
SkyNet & 0.390 & 0.442 & 0.158 & 30.15\\
\bottomrule
\end{tabular}
\end{table}

\vspace{-5pt}
\section{Conclusions}
\vspace{-5pt}
In this paper, we proposed SkyNet, as well as a hardware-efficient method to generate compact DNNs for object detection running on embedded GPUs and embedded FPGAs. SkyNet design methodology is a novel bottom-up DNN design flow which can capture hardware limitations using realistic hardware feedbacks and deliver DNNs with great balance between software and hardware metrics, such as DNN inference accuracy and throughput performance. SkyNet was demonstrated on the 56th IEEE/ACM DAC-SDC low power object detection challenge and won the first place winner award for both GPU and FPGA tracks. We also extended SkyNet to handle object tracking task and it delivered 1.60X and 1.73X higher FPS, and 37.20X smaller parameter size with comparable accuracy when compared to the state-of-the-art Siamese trackers with ResNet-50 backbone. 

\vspace{-5pt}
\section{Acknowledgments}
\vspace{-5pt}
This work was partly supported by the IBM-Illinois Center for Cognitive Computing System Research (C$^3$SR) -- a research collaboration as part of IBM AI Horizons Network. 




\bibliography{example_paper}
\bibliographystyle{sysml2019}



\end{document}